\documentclass[10pt]{article}
\usepackage[letterpaper]{geometry}
\usepackage{hicss}
\usepackage{times}
\usepackage[none]{hyphenat}
\usepackage{url}
\usepackage{latexsym}
\usepackage{minted}
\usepackage{indentfirst}
\usepackage{graphicx}
\graphicspath{{images/}}
\usepackage[
    style=apa,
  ]{biblatex}
\usepackage{caption}
\usepackage{array}
\usepackage{subcaption}
\usepackage{pifont}
\usepackage{xcolor}
\usepackage{listings}
\usepackage{multirow} 
\usepackage{amssymb}
\usepackage{amsmath}
\usepackage{booktabs}
\usepackage{threeparttable}
\usepackage{microtype}
\usepackage{helvet} 
\usepackage{caption}
\usepackage{subcaption}

\DeclareCaptionLabelSeparator{hicss-double}{.~ }

\newcommand{\greencheck}{\textcolor{green!60!black}{\checkmark}}
\newcommand{\redx}{\textcolor{red}{\text{\sffamily X}}}

\title{Evaluating the Semantic-to-Geometric Gap in Adversarial Defenses Against Vision-Language Model-Based Plagiarism}

\author{Christopher Burger \\
 \small{Pelican Quantitative} \\
 {chris@pelicanqc.com} \\ \\\And
 Christina Trotter \\
\small{The University of Mississippi} \\
 {cjtrotte@olemiss.edu } \\ \And
  Joseph Carlisle \\
 \small{The University of Mississippi} \\
 {jcarlis1@olemiss.edu} \\ \And
 Charles Walter\\
 \small{The University of Mississippi} \\
 {cwwalter@olemiss.edu} \\ }

\date{}

\begin{document}
\maketitle
\begin{abstract}

The rapidly advancing capabilities of vision-language models (VLMs) present a systemic challenge to academic integrity. VLMs now allow students to bypass meaningful engagement by capturing and submitting graphical problems as singular images, a practice we define as ``trivial plagiarism." To provide educators with actionable data on VLM limitations, we investigate the efficacy of heuristic adversarial image transformations designed to degrade model performance while remaining human-interpretable. Through a two-phase evaluation of introductory assessments, we manually assess baseline VLM performance on circuit diagrams, followed by an automated large-scale evaluation of topological structures (logic gates) and coordinate geometry (Karnaugh maps). We find that while highly capable VLMs can exhibit appreciable robustness, all models suffer vulnerability to adversarial perturbations. We conclude that while visual perturbations act as a viable near-term stopgap, long-term assessment security requires educators to reapproach assessment design given continually increasing VLM performance.

\end{abstract}
\subsubsection*{Keywords:}

Vision-Language Models, Academic Integrity, Adversarial Examples, Assessment Design
\section{Introduction}

The integration of vision-language models (VLMs) into academic settings has created a tension between their potential as powerful learning aids and their use as tools for academic dishonesty. While text-based plagiarism via large-language models (LLMs) is a recognized issue \parencite{toba2023inappropriatebenefitsidentificationchatgpt,Cotton03032024,wahle-etal-2022-large,teel}, the recent introduction of multimodal models capable of interpreting images represents a substantially lower barrier for plagiarism \parencite{miles2022students,peters2025redefining}. This capability allows students to circumvent the learning process entirely by submitting a raw image of a problem, a method we define as \textit{trivial plagiarism}. Trivial plagiarism requires minimal effort by the student,  who no longer needs to understand (at least some aspect of) the problem to obtain a solution through appropriately crafted prompts. Rather, the student needs only take a picture of the problem (or wear smart glasses with cameras recording) to obtain a solution, not thinking or understanding any aspect of the problem or solution.

Traditional plagiarism required some understanding of the content being assessed to remain undetected \parencite{peters2025redefining}. Previously, obtaining correct results from an LLM often required careful prompt engineering, a task that necessitated a degree of patience and some level of familiarity with the subject matter. This interaction, while potentially aimed at academic misconduct, created a soft barrier that could still allow for some incidental learning. However, the effectiveness of current VLM models on direct image inputs has largely removed this barrier. A student can now upload a photograph or screenshot of a graphical problem, from a circuit diagram to a chemical equation, and receive a correct solution with no real intellectual engagement. This shift separates traditional plagiarism from trivial plagiarism, as there is no understanding from the student but detection based only on the answer is difficult or impossible.

Our work explores a defensive strategy against this form of plagiarism by leveraging adversarial examples. While prior research on adversarial defenses in the context of academic integrity has focused predominantly on text inputs \parencite{salim2024impedingllmassistedcheatingintroductory}, image-based inputs remain a relatively unexplored domain. We argue that image perturbations represent a more robust and efficient defensive mechanism. Unlike perturbations in natural language, which must carefully preserve syntactic and semantic meaning to remain coherent \parencite{burger2023explanations}, image transformations can tolerate a greater degree of alteration before a human observer's understanding of the content is compromised \parencite{3327757.3327854,captcha,marrVision}.

The goal of this paper is to evaluate the plausibility of computationally efficient image transformations as a deterrent to trivial plagiarism, and to investigate the limitations of VLMs when processing visual assessments. We present a two-phase evaluation. Phase 1 manually evaluates a dataset of introductory circuit analysis problems containing both text and diagrams to establish the baseline effectiveness of heuristic visual obfuscation (e.g., pixelated noise and affine warping). Phase 2 introduces a fully automated pipeline evaluating purely visual topological structures (logic diagrams) and geometric structures (Karnaugh maps). This automated approach allows for extended and repeated testing to confirm model efficacy given that there exists unavoidable randomness when using modern LLMs. Ultimately, we argue that while visual perturbations act as a viable near-term stop-gap by degrading model accuracy, the long-term solution for assessment security requires exploiting these underlying spatial and geometric blind spots rather than engaging in a futile arms race of image degradation.


\section{Background \& Related Work}

The integration of LLMs into educational environments has invoked a significant debate regarding academic integrity and the validity of traditional assessments \parencite{Nikolic02072024,Nikolic02072024,Cotton03032024,app14104115}. While LLMs undoubtedly offer benefits, especially through asynchronicity and personalization of learning, the question of how to prevent the circumvention of assessments becomes more prominent as LLMs transition beyond text into multimodal VLMs, which are capable of accepting a growing variety of inputs. Attempting to algorithmically detect LLM-generated output has proven systematically unreliable; recent studies demonstrate that AI detectors yield appreciable false positive rates and can fail to catch sophisticated AI use \parencite{Weber_Wulff_2023, ofqual2024,burger2025,GPT4AcademicDetection}. This detection collapse necessitates a pivot toward proactive, input-level defenses.

In an attempt to maintain academic integrity, defense mechanisms involving adversarial perturbations (a known weakness of many classic deep learning models) have been developed. The closest directly related to our work being \textcite{salim2024impedingllmassistedcheatingintroductory}, which focuses on text-based perturbation to corrupt code generation for programming assignments. However, we seek to deal with VLMs and image-based input. Much of the adversarial machine learning literature focused on images has predominantly utilized gradient-based attacks, such as the Fast Gradient Sign Method \parencite{goodfellow2014explaining}. However, these attacks require full ``white-box" access, meaning the attacker needs complete visibility and control over the model's proprietary architecture, internal weights, and gradients. Because students rely on closed-source, black-box platforms (e.g., ChatGPT), executing white-box attacks is impossible for educators, as those resources are exclusively held by the parent tech corporations. Even if such access were available, mathematically complex gradient attacks remain far too computationally expensive to be integrated into high-volume educational content management systems. 

Vision transformers, which power modern state-of-the-art VLMs, have also demonstrated inherent robustness against traditional black-box adversarial attacks \parencite{mahmood2021robustness}, necessitating an even greater computational budget for classic adversarial methods. Consequently, instead of relying on computationally prohibitive deep-learning exploits, we investigate heuristic, visually perceptible perturbations based on natural forms of image degradation.

Our defense strategy is informed by recent findings regarding the ``cognitive'' and spatial limitations of VLMs. While vision-language models have what amounts to a holistic comprehension of visual environments, recent benchmarks reveal deficiencies in genuine spatial reasoning; models suffer from a fundamental ``semantic-to-geometric gap'' where qualitative object recognition fails to translate into high-fidelity geometric computation. Specifically, VLMs perform adequately on static spatial tasks or basic topological connections, but their performance drops appreciably on tasks requiring dynamic transformations, mental rotation, or strict coordinate geometry \parencite{stogiannidis2025mindgapbenchmarkingspatial, deng2025internspatialcomprehensivedatasetspatial}. By targeting this spatial weakness with heuristic visual obfuscation, we aim to bridge the gap between adversarial machine learning and practical assessment design. We acknowledge that these perturbations are fundamentally crude; they represent a necessary trade-off: optimizing for rapid, automated pipeline integration rather than effectively human undetectable perturbations. Ultimately, while this visual disruption serves as an effective near-term stop-gap, the long-term assurance of academic integrity will likely require educators to rethink assessment redesign rather than simply obscuring the input.




\begin{center}
\begin{figure*}[ht]
    
    \centering
    \begin{subfigure}[t]{0.32\textwidth}
        \includegraphics[width=\linewidth,height=0.75\linewidth]{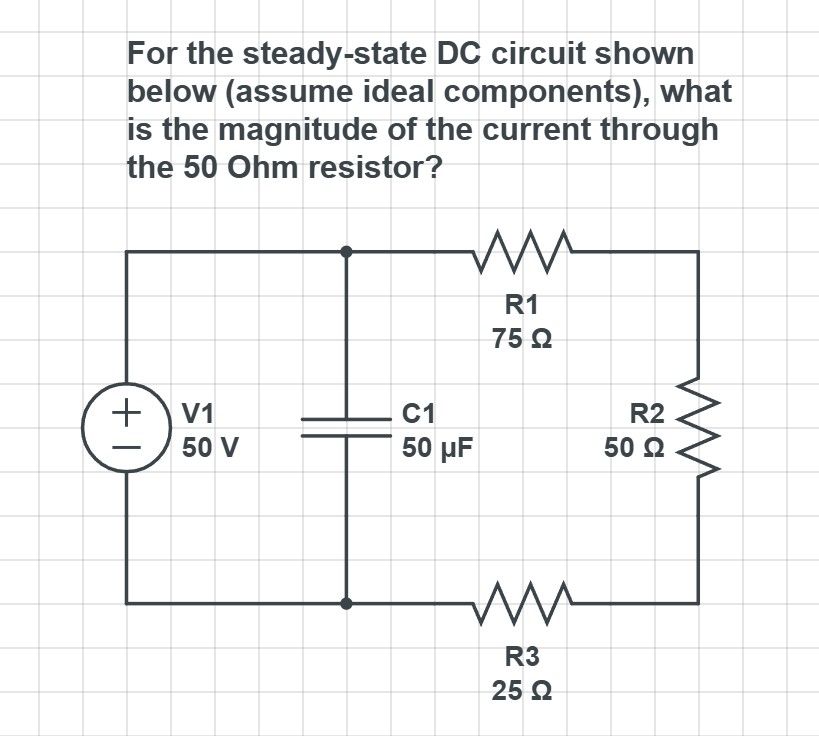}
        \caption{Original, unperturbed image.}
    \end{subfigure}
    \hfill
    \begin{subfigure}[t]{0.32\textwidth}
        \includegraphics[width=\linewidth,height=0.75\linewidth]{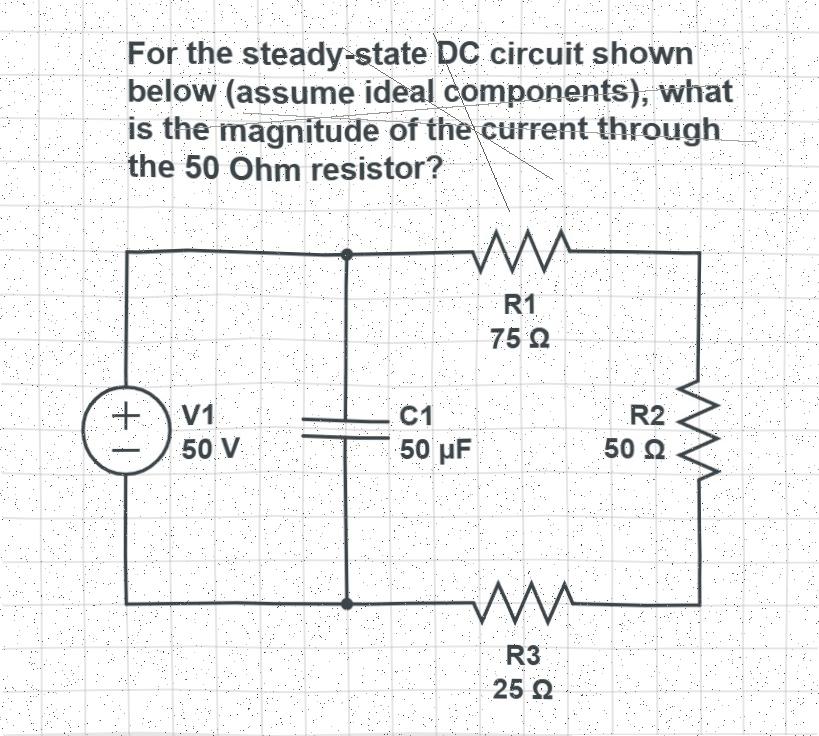}
        \caption{Perturbed but legible.}
    \end{subfigure}
    \hfill
    \begin{subfigure}[t]{0.32\textwidth}
        \includegraphics[width=\linewidth,height=0.75\linewidth]{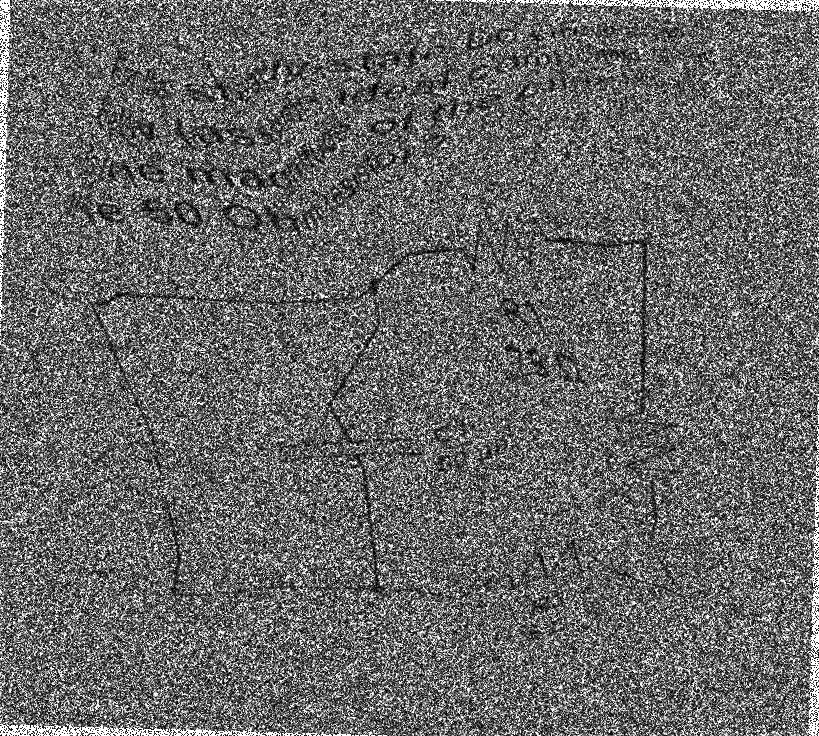}
        \caption{Perturbed and illegible.}
        \label{fig:degraded}
    \end{subfigure}
    \caption{Legibility comparison between perturbed problem images and the original.}
    \label{fig:perturbation_comparison}
\end{figure*}
\end{center}





\begin{figure*}[ht]
    \centering

    \begin{subfigure}[t]{0.32\textwidth}
        \includegraphics[width=\linewidth,height=0.75\linewidth]{RC_Circuit.jpg}
        \caption{Original}
    \end{subfigure}
    \hfill
    \begin{subfigure}[t]{0.32\textwidth}
        \includegraphics[width=\linewidth,height=0.75\linewidth]{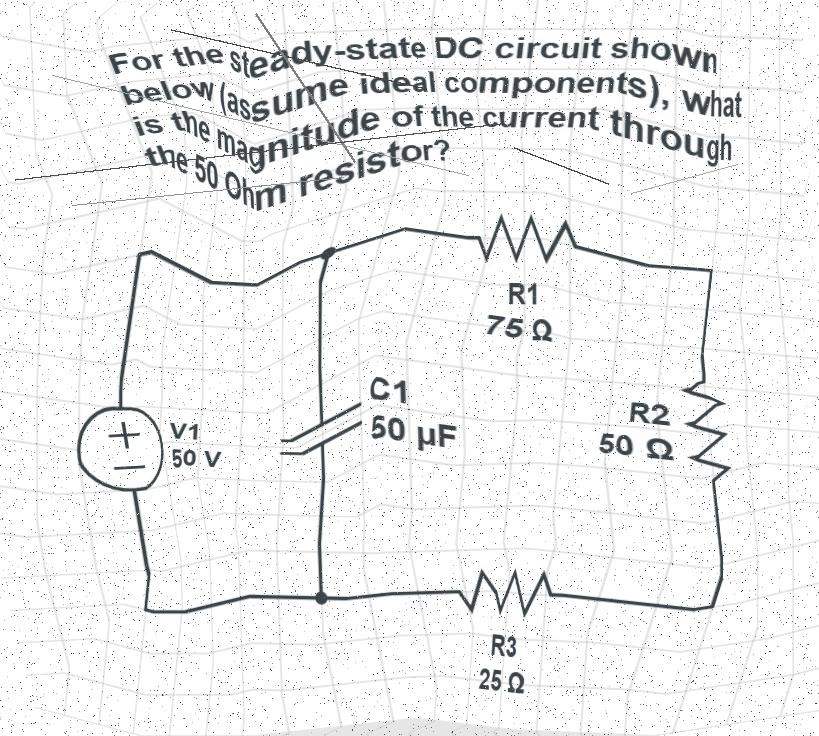}
        \caption{Noise + Lines + Warping}
    \end{subfigure}
    \hfill
    \begin{subfigure}[t]{0.32\textwidth}
        \includegraphics[width=\linewidth,height=0.75\linewidth]{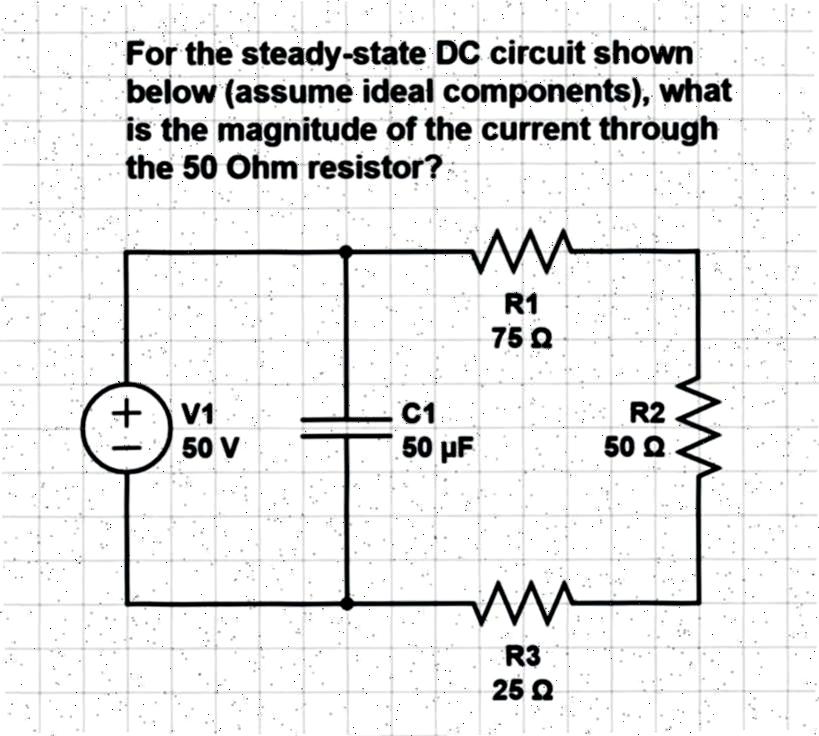}
        \caption{Photocopy}
    \end{subfigure}


    \begin{subfigure}[t]{0.32\textwidth}
        \includegraphics[width=\linewidth,height=0.75\linewidth]{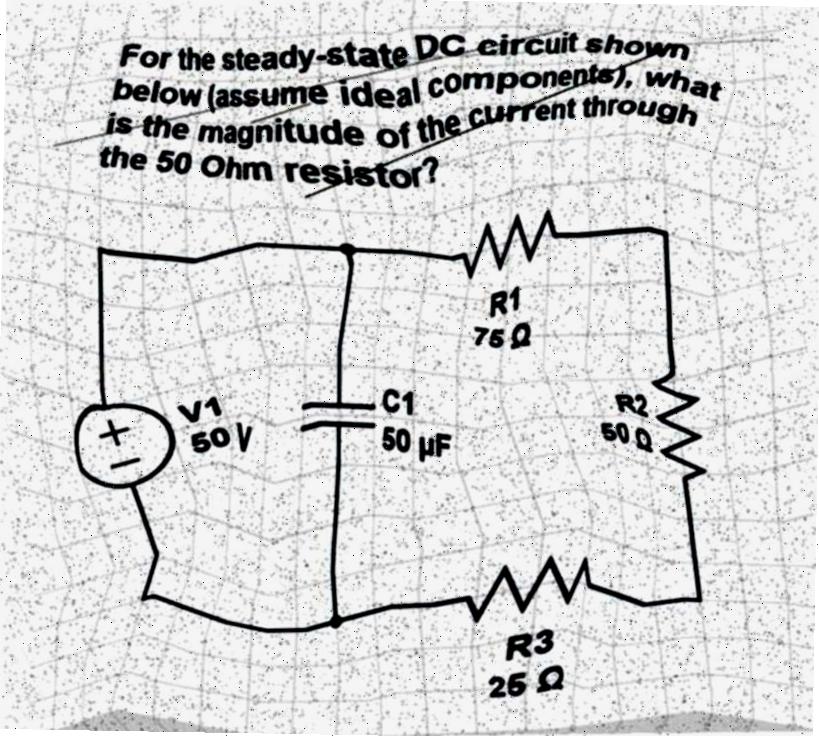}
        \caption{All perturbations on full image}
    \end{subfigure}
    \hfill
    \begin{subfigure}[t]{0.32\textwidth}
        \includegraphics[width=\linewidth,height=0.75\linewidth]{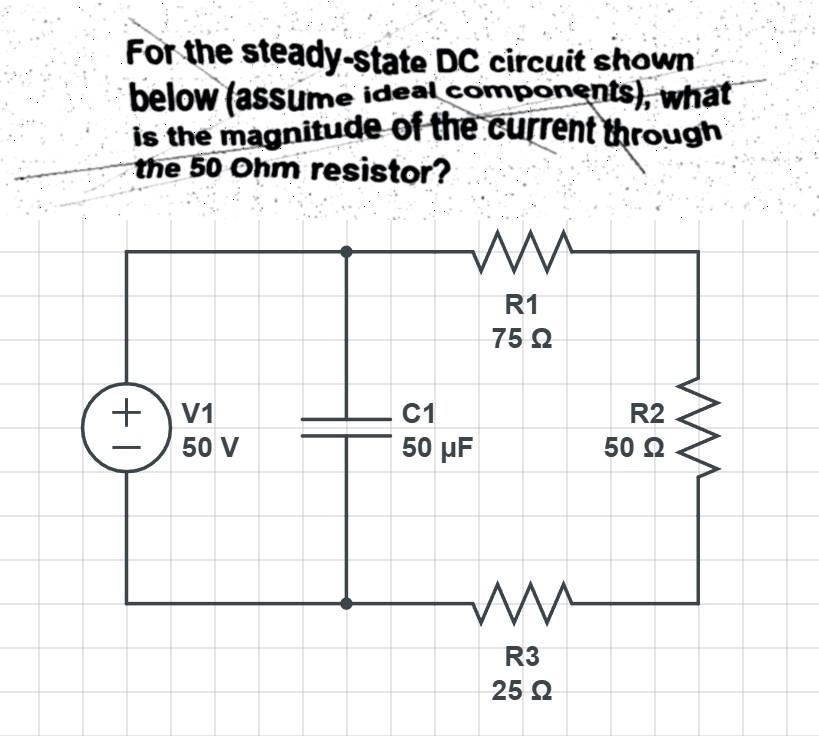}
        \caption{All perturbations on text prompt only}
    \end{subfigure}
    \hfill
    \begin{subfigure}[t]{0.32\textwidth}
        \includegraphics[width=\linewidth,height=0.75\linewidth]{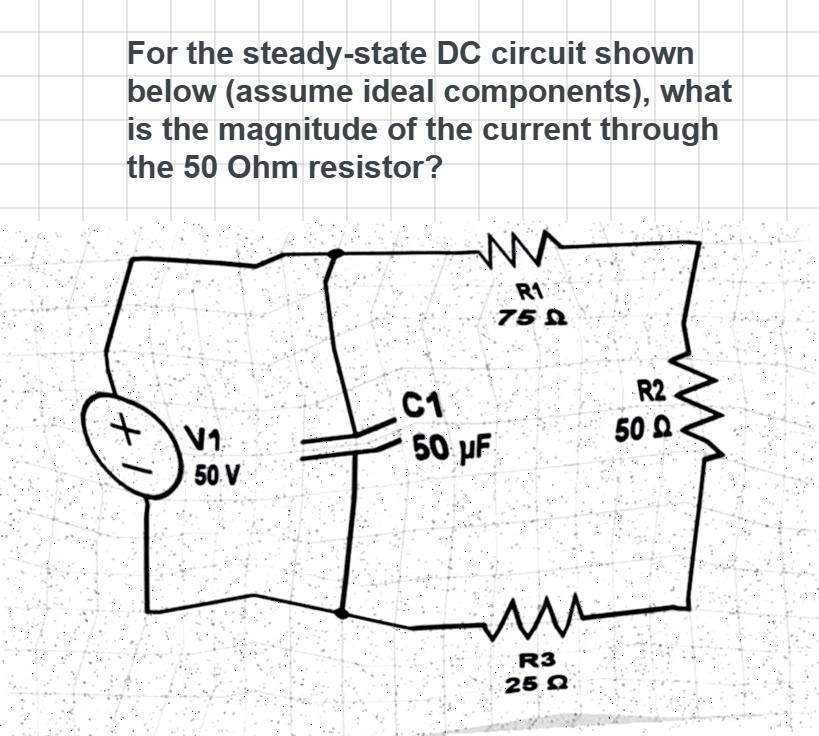}
        \caption{All perturbations on figure only}
    \end{subfigure}

    \caption{Perturbations and their valid regions on the image. Note that the intensity of the perturbations here is deliberately high to emphasize their effect. Subfigures (b) and (d) are pushing the boundary of what we would consider to be reasonably comprehensible.}
    \label{fig:example1}
\end{figure*}

\vspace{-20pt}

\section{Methodology}
\label{sec:criteria}

To be practically useful in an educational setting, an adversarial defense mechanism must satisfy two criteria. First, it must be computationally efficient. We avoid strictly defining efficient, but seek methods with generation times that would not feel burdensome if integrated in a content management system. For example, a few minutes to generate quality perturbations with an acceptable likelihood of defeating VLM models may be considered reasonable, whereas hours of computation is not. Second, it must maintain human interpretability; the perturbations must degrade the model's performance without altering the fundamental problem or burdening the student with excessive artifacts. We do not seek to find mathematically optimal, white-box adversarial noise, but rather to determine if fast, heuristic methods can induce erroneous VLM outputs while keeping the perturbed document legible.

\subsection{Perturbation Methods}\label{sec:perturbation_methods}
The methods used in the perturbation process were chosen to mimic common causes of image degradation. The goal here is not to find an ``optimal'' perturbation from an iterative process, but instead to determine if fast and efficient methods can create what looks to be ``natural'' (or, at least, human interpretable) imperfections in an image representation of a problem that induce erroneous output from a VLM.


\noindent We utilize three main perturbations that emulate common negative conditions when reproducing an image. These perturbations are:  \textbf{Warping},  \textbf{Pixel Noise}, and \textbf{Lines} with random start and end points. When we use the term ``random'', we mean the values are generated arbitrarily and uniformly. Additionally, we introduce a multi-perturbation method to simulate a \textbf{Photocopy} as a perturbation that may result from normal use of material.

\noindent \textbf{Warping:} Used to simulate physical paper distortion (like crinkling, folding, etc). We apply a piecewise affine transformation by first applying a grid of control points over the image. Then, the locations of these points are shifted randomly, 
with a defined maximum displacement. A piecewise affine transform function then calculates the distortion needed to move the grid points from the original to the new location. Ultimately, this results in stretching or compressing different parts of the image locally.

\noindent \textbf{Pixel Noise:} Used to simulate imaging sensor noise, dust / debris, or print imperfections. Some proportion of pixels are randomly selected across the image for perturbation. Each selected pixel is then filled with a randomly chosen shade of grey.

\noindent \textbf{Lines:} Used to simulate scratches on lenses or scanner glass, fold lines or creases in paper, streaks from faulty copier drums or sensor elements, etc. We generate a small, random number of straight lines (3 to 7), with random start and end points, random shade of gray, and a random small width (1 or 2 pixels). The line perturbations are applied only to the problem text and never the figure as the addition of a line can fundamentally alter the structure of the circuit.

\noindent \textbf{Photocopy:} Used to emulate the effect of degradation from the digitization/image conversion process when a document is repeatedly copied, like say a photocopy of a photocopy, or a picture of a photocopy. This method has seven stages applied in sequence and then the result is fed back into the method for some number of iterations to simulate the progressive degradation.
\begin{enumerate}
    \item \textbf{Greyscale conversion:} Converts the image back to greyscale to disable any unexpected color additions from prior perturbations.
    \item \textbf{Contrast adjustment:} Randomly increases the contrast by a small factor. Useful for simulating loss of detail from digitization, e.g., `crushed blacks', or `blown-out whites'. Default parameter set to the range of 1.1 to 1.4.
    \item \textbf{Brightness adjustment:} Random increases or decreases the brightness by a small factor. Helps simulate variations in exposure settings or lamp age for a scanner / camera. Default parameter set to the range of 0.95 to 1.05.
    \item \textbf{Blurring:} A Gaussian blur filter is applied with a small, random radius. Used to provide a small, cumulative loss of sharpness and fine detail. Default parameter set to range 0.3 to 0.8.
    \item \textbf{Skew:} Applied a slight rotation (up to 1 degree). Used to mimic misalignment. Default parameter set to active.
    \item \textbf{Pixel noise:} Functionally equivalent to the pixel noise perturbation described above. Default parameter set to active.
    \item \textbf{Compression artifacts:} Performs a save to a temporary JPEG file with a random quality level (50 to 90), the file is then reloaded and used for future iterations or permanently saved. Used to simulate the degradation of repeated digitization in lossy formats. Default parameter range set to 50\% to 90\%.
\end{enumerate}

\noindent Finally, we include an option to control where the above perturbations are applied. By default the entire image is perturbed, but we allow for the perturbation of only the question text or the associated figure. 

While the effectiveness of the perturbations is clearly important, this work is concentrated on exploring if naive selections of the above perturbations demonstrate enough efficacy to justify the time and computational expense to optimize this process. We note that with the above perturbation methods all of our practicality conditions are met. First, the code needed to generate these perturbations is simple with well specified parameters that will allow future tuning of the perturbations. And second, the process is fast in generating and applying these perturbations. It takes only fractions of a second per image on a modest desktop computer for all options enabled at any reasonable value (that is, values that do not degrade the image to the point of being unrecognizable).





\subsection{Phase 1: Manual Evaluation (Circuit Diagrams)} Our evaluation is divided into two distinct phases. Phase 1 serves as an initial heuristic baseline, evaluating problems that contain both necessary textual components and visual diagrams. We selected introductory DC circuit analysis as the domain, curating a dataset of instructor resources with provided (and verified) solutions where a short text prompt (10-20 words) and a simple schematic (restricted to resistor networks) are both required to solve for a specific value (e.g., current or voltage). Our reasoning behind the choice of DC circuit problems is to choose subject matter that involves both text and graphics simultaneously while being challenging enough (to VLMs) to increase the likelihood of efficacy of relatively minor perturbations (and so hopefully less disruptive to viewers).

This phase utilizes a manual evaluation process via the models' standard web interfaces to simulate the exact workflow of a cheating student. We apply the perturbations (Section \ref{sec:criteria}) selectively: to the text only, to the figure only, or to the entire combined image, allowing us to isolate which region the model relies on most heavily. Because this process relies on both manual web-interface input and processing of the results, the sample size is constrained, but it provides a necessary qualitative baseline for real-world VLM interaction.
\\

\noindent The evaluation process was performed as follows:

\begin{enumerate}
    \item All prompts used the identical structure: \textit{\begin{quote}
        Can you solve the problem in the attached image?
    \end{quote}}
    \item Errors in processing (e.g. network timeouts) resulted in the creation of a new conversation as to not bias future output.
    \item Complete output of all responses was recorded and compared to known correct solutions from guaranteed human generated content.
\end{enumerate}


\subsection{Phase 2: Automated Evaluation (Topological and Geometric Rigidity)} Phase 2 transitions to a fully automated pipeline designed to evaluate the limits of VLM spatial reasoning at scale. Unlike Phase 1, the problem text is provided directly in the system prompt (which remains identical for each distinct problem type), and so the adversarial perturbations are applied only to the image. This phase reduces the stochasticity of web interfaces by utilizing direct API access, fixing the generation temperature to $0.0$, and evaluating each image three separate times to generate a robust ``Pass@3'' majority-vote metric.
\\

\noindent To ensure consistency and correct evaluation, the automated prompting and evaluation pipeline is as follows:
\\

    \noindent The models were instructed using a highly constrained zero-shot template. To prevent standard textbook syntax from interfering with evaluation, the prompt demanded a specific output format (FINAL ANSWER: $<$expression$>$) and enforced explicit negative constraints (e.g., instructing the model to use \textasciitilde \; for NOT).
    
    \noindent Degraded images were passed directly to the model APIs alongside the system prompt. To prevent data loss or false negatives caused by server-side stochasticity, network timeouts and rate-limit errors (e.g., HTTP 429) triggered an pause-and-retry loop.
    
    \noindent The raw text output from the VLM was parsed using regular expressions to isolate the final answer block. A programmatic string-cleaning safety net was applied to standardize common LLM formatting inconsistencies that bypassed prompt constraints (e.g., automatically replacing a rogue $+$ sign with the requested $|$ operator for a logical OR).
     
   \noindent Rather than relying on brittle, exact-string matching, the sanitized boolean expression was passed into SymPy, a Python library for symbolic mathematics. The model's logic was evaluated against the known ground-truth expression using a strict symbolic equivalence check. This ensured that functionally identical answers written in varying associative orders (e.g., $A \ \& \ B$ versus $B \ \& \ A$) were correctly counted as equivalent.

\begin{figure}[t]
    \centering

    \begin{subfigure}{0.48\columnwidth}
        \centering
        \includegraphics[width=\linewidth]{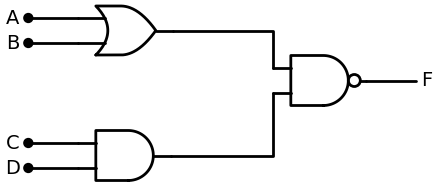}
        \caption{Logic Diagram}
        \label{fig:a}
    \end{subfigure}
    \hfill
    \begin{subfigure}{0.48\columnwidth}
        \centering
        \includegraphics[width=\linewidth]{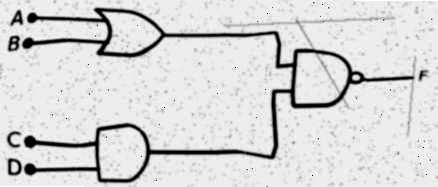}
        \caption{Perturbed Figure \ref{fig:a}}
        \label{fig:b}
    \end{subfigure}

    \vspace{0.5em}

    \begin{subfigure}{0.48\columnwidth}
        \centering
        \includegraphics[width=\linewidth]{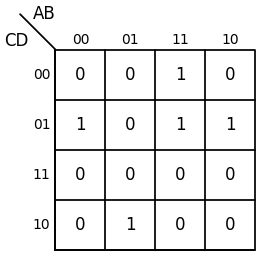}
        \caption{Karnaugh Map}
        \label{fig:c}
    \end{subfigure}
    \hfill
    \begin{subfigure}{0.48\columnwidth}
        \centering
        \includegraphics[width=\linewidth]{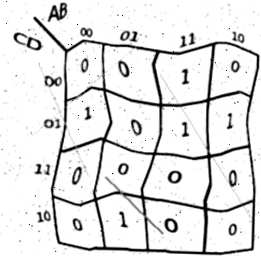}
        \caption{Perturbed Figure \ref{fig:c}}
        \label{fig:d}
    \end{subfigure}

    \caption{Phase 2 automated evaluation datasets, contrasting continuous topological routing (a, b) with geometric alignment (c, d) under baseline and perturbed conditions.}
    \label{fig:grid}
\end{figure}

\noindent Phase 2 then tests two distinct visual structures:

\noindent \textbf{Topological Routing (Digital Logic Circuits):} A dataset of mathematically unique logic gate diagrams (e.g., AND, OR, NOT cascades) (Figure \ref{fig:grid}). This tests whether models can trace continuous paths (edges) between nodes, even when lines are broken by pixel noise or affine warping.

\noindent \textbf{Coordinate Geometry (Karnaugh Maps):} A dataset of 4-variable K-Maps, including both randomized distributions and specific pedagogical edge cases (e.g., four-corner wrap-around groupings) (Figure \ref{fig:grid}). This tests the models' reliance on  spatial-coordinate rigidity, as minor visual warping can shift binary values across rigid cell boundaries. 

\subsection{Model Selection} The selection of VLMs is dictated by accessibility; we evaluate models comparable to those available to standard undergraduate students, bypassing specialized or open-source models that require local deployment. For the manual evaluation in Phase 1, we utilize the flagship models available via free or university-provided web interfaces: Google's Gemini 2.5 Pro and Gemini 2.5 Flash, Anthropic's Claude Sonnet 4, and OpenAI's GPT-4 Turbo (via Microsoft Copilot). For the automated evaluation in Phase 2, we target the  ``lightweight flagship'' tier via developer APIs. We utilize Gemini 2.5 Flash, Gemini 2.5 Pro, OpenAI's GPT-4o-mini, and Anthropic's Claude 4.5 Haiku to provide the closest match to the models used in Phase 1. Additionally, these models represent a balance between capability, speed, and scope of deployment (that is, models commonly used by third parties as the backend for their services) for multimodal architectures currently available to the general public.

\section{Results}
Our evaluation is structured around the two experimental phases. Phase 1 details the manual evaluation of regional perturbations on introductory circuit analysis problems. Phase 2 presents the automated evaluation of specific spatial and geometric vulnerabilities.

\subsection{Phase 1: Manual Evaluation}
To establish a baseline for real-world VLM interaction, we manually evaluated a curated subset of 25 introductory circuit problems. Authors ensured that each perturbed image tested remained legible enough to complete the problem successfully. From this baseline, a representative sample of seven images was selected to apply the heuristic transformations. This subset was chosen to encompass a range of difficulties, with the difficulty increasing by question number.

Table \ref{tab:unified_adversarial_results} summarizes the efficacy of perturbations across the question text, the figure, and the full combined image. We define a \textit{successful perturbation} as one that alters the output of the model, a \textit{positive perturbation} as a shift from an incorrect to a correct answer, and a \textit{negative perturbation} as a shift from correct to incorrect. In general, the perturbation process produced some level of effectiveness across all model types, with certain models being more susceptible to specific perturbation types over others. For example, Claude was most susceptible to the photocopy perturbation alone, with enabling the remaining noise, lines, and warping perturbations actually reducing overall success. Gemini as a whole was more robust, with 2.5 Pro maintaining strong, but not complete, resistance to adversarial perturbation. The most interesting finding was the presence of positive perturbations, but due to the stochasticity of the web-interface and the time requirements for manually performing the experimentation, this cannot be proven to be an artifact of the perturbation process itself.

To mitigate the issue of stochasticity observed in the web-interface trials, we constructed an automated pipeline allowing us to perform a much larger, repeated set of experiments. This necessitated a transition from standard circuit diagrams as algorithmically generating complex, readable circuit diagrams with verified ground-truth solutions at scale was infeasible since it requires extensive manual verification. By shifting to well-defined mathematical structures like logic gates and coordinate maps, we could fully automate the generation and evaluation loop, enabling larger samples sizes to more confidently measure model robustness.

\begin{table*}[t]
\centering
\begin{threeparttable}
    \caption{Comparison of Model Robustness Under Text, Figure or Both (Full) Perturbations}
    \label{tab:unified_adversarial_results}
    \centering
    \footnotesize
    \setlength{\tabcolsep}{5.25pt} 
    \begin{tabular}{ll cccc cccc cccc cccc}
        \toprule
        & & \multicolumn{4}{c}{\textbf{Claude Sonnet 4}} & \multicolumn{4}{c}{\textbf{Gemini 2.5 Pro}} & \multicolumn{4}{c}{\textbf{Gemini 2.5 Flash}} & \multicolumn{4}{c}{\textbf{GPT-4}} \\
        \cmidrule(lr){3-6} \cmidrule(lr){7-10} \cmidrule(lr){11-14} \cmidrule(lr){15-18}
        \textbf{Config} & \textbf{Q} & \textbf{Base} & \textbf{Txt} & \textbf{Fig} & \textbf{Full} & \textbf{Base} & \textbf{Txt} & \textbf{Fig} & \textbf{Full} & \textbf{Base} & \textbf{Txt} & \textbf{Fig} & \textbf{Full} & \textbf{Base} & \textbf{Txt} & \textbf{Fig} & \textbf{Full} \\
        \midrule

        \multirow{7}{*}{\textbf{All}} 
        & 1  & \redx & \redx & \redx & \redx & \redx & \redx & \greencheck & \greencheck & \redx & \redx & \redx & \redx & \redx & \redx & \redx & \redx \\
        & 5  & \greencheck & \redx & \redx & \redx & \redx & \redx & \redx & \redx & \greencheck & \redx & \redx & \redx & \redx & \greencheck & \redx & \redx \\
        & 6  & \greencheck & \greencheck & \greencheck & \greencheck & \greencheck & \greencheck & \greencheck & \greencheck & \greencheck & \greencheck & \greencheck & \greencheck & \greencheck & \greencheck & \greencheck & \greencheck \\
        & 7  & \greencheck & \greencheck & \greencheck & \greencheck & \greencheck & \greencheck & \greencheck & \greencheck & \greencheck & \greencheck & \greencheck & \greencheck & \greencheck & \greencheck & \greencheck & \greencheck \\
        & 12 & \greencheck & \redx & \greencheck & \greencheck & \greencheck & \greencheck & \greencheck & \greencheck & \greencheck & \greencheck & \greencheck & \greencheck & \greencheck & \greencheck & \redx & \greencheck \\
        & 13 & \greencheck & \greencheck & \greencheck & \greencheck & \greencheck & \greencheck & \greencheck & \greencheck & \greencheck & \greencheck & \greencheck & \greencheck & \greencheck & \greencheck & \greencheck & \greencheck \\
        & 14 & \greencheck & \greencheck & \redx & \greencheck & \greencheck & \greencheck & \greencheck & \greencheck & \greencheck & \greencheck & \redx & \redx & \redx & \redx & \redx & \redx \\
        \cmidrule(lr){2-18}
        & \textbf{Acc} & \textbf{6/7} & \textbf{4/7} & \textbf{4/7} & \textbf{5/7} & \textbf{5/7} & \textbf{5/7} & \textbf{6/7} & \textbf{6/7} & \textbf{6/7} & \textbf{5/7} & \textbf{5/7} & \textbf{5/7} & \textbf{4/7} & \textbf{5/7} & \textbf{3/7} & \textbf{4/7} \\
        \midrule

        \multirow{7}{*}{\textbf{Photo}} 
        & 1  & \redx & \redx & \redx & \redx & \redx & \greencheck & \redx & \redx & \redx & \redx & \greencheck & \redx & \redx & \redx & \redx & \redx \\
        & 5  & \greencheck & \redx & \redx & \redx & \redx & \greencheck & \greencheck & \greencheck & \greencheck & \redx & \redx & \redx & \redx & \greencheck & \redx & \greencheck \\
        & 6  & \greencheck & \greencheck & \greencheck & \greencheck & \greencheck & \greencheck & \greencheck & \greencheck & \greencheck & \greencheck & \greencheck & \greencheck & \greencheck & \greencheck & \greencheck & \greencheck \\
        & 7  & \greencheck & \greencheck & \greencheck & \greencheck & \greencheck & \greencheck & \greencheck & \greencheck & \greencheck & \greencheck & \greencheck & \greencheck & \greencheck & \greencheck & \greencheck & \greencheck \\
        & 12 & \greencheck & \redx & \redx & \redx & \greencheck & \greencheck & \greencheck & \greencheck & \greencheck & \greencheck & \greencheck & \greencheck & \greencheck & \greencheck & \greencheck & \greencheck \\
        & 13 & \greencheck & \greencheck & \redx & \redx & \greencheck & \greencheck & \greencheck & \greencheck & \greencheck & \greencheck & \greencheck & \greencheck & \greencheck & \greencheck & \greencheck & \greencheck \\
        & 14 & \greencheck & \greencheck & \redx & \greencheck & \greencheck & \greencheck & \greencheck & \greencheck & \greencheck & \redx & \redx & \greencheck & \redx & \greencheck & \greencheck & \redx \\
        \cmidrule(lr){2-18}
        & \textbf{Acc} & \textbf{6/7} & \textbf{4/7} & \textbf{2/7} & \textbf{3/7} & \textbf{5/7} & \textbf{7/7} & \textbf{6/7} & \textbf{6/7} & \textbf{6/7} & \textbf{4/7} & \textbf{5/7} & \textbf{5/7} & \textbf{4/7} & \textbf{6/7} & \textbf{5/7} & \textbf{5/7} \\
        \midrule

        \multirow{7}{*}{\textbf{NLW}} 
        & 1  & \redx & \redx & \redx & \redx & \redx & \greencheck & \greencheck & \redx & \redx & \greencheck & \redx & \greencheck & \redx & \redx & \redx & \redx \\
        & 5  & \greencheck & \greencheck & \greencheck & \redx & \redx & \greencheck & \redx & \redx & \greencheck & \redx & \redx & \redx & \redx & \greencheck & \greencheck & \greencheck \\
        & 6  & \greencheck & \greencheck & \greencheck & \greencheck & \greencheck & \greencheck & \greencheck & \greencheck & \greencheck & \greencheck & \greencheck & \greencheck & \greencheck & \greencheck & \greencheck & \greencheck \\
        & 7  & \greencheck & \greencheck & \greencheck & \greencheck & \greencheck & \greencheck & \greencheck & \greencheck & \greencheck & \greencheck & \greencheck & \greencheck & \greencheck & \greencheck & \greencheck & \greencheck \\
        & 12 & \greencheck & \greencheck & \greencheck & \greencheck & \greencheck & \greencheck & \greencheck & \greencheck & \greencheck & \greencheck & \greencheck & \greencheck & \greencheck & \greencheck & \greencheck & \greencheck \\
        & 13 & \greencheck & \greencheck & \greencheck & \greencheck & \greencheck & \greencheck & \greencheck & \greencheck & \greencheck & \greencheck & \greencheck & \greencheck & \greencheck & \greencheck & \greencheck & \redx \\
        & 14 & \greencheck & \greencheck & \greencheck & \greencheck & \greencheck & \greencheck & \greencheck & \redx & \greencheck & \greencheck & \redx & \greencheck & \redx & \redx & \redx & \redx \\
        \cmidrule(lr){2-18}
        & \textbf{Acc} & \textbf{6/7} & \textbf{6/7} & \textbf{6/7} & \textbf{5/7} & \textbf{5/7} & \textbf{7/7} & \textbf{6/7} & \textbf{4/7} & \textbf{6/7} & \textbf{6/7} & \textbf{4/7} & \textbf{6/7} & \textbf{4/7} & \textbf{5/7} & \textbf{5/7} & \textbf{4/7} \\
        \bottomrule
    \end{tabular}
    \begin{tablenotes}
        \small
        \item \textit{Note:} \textbf{Base} denotes baseline model performance (no perturbations). \textbf{Txt} and \textbf{Fig} denote performance under text prompts and image modifications respectively. \textbf{NLW} denotes the combination of the Noise, Lines and Warping perturbations. Cases where the model recovers from a baseline failure ($\redx \rightarrow \greencheck$) indicate counter-intuitive helper perturbations.
    \end{tablenotes}
\end{threeparttable}
\end{table*}

\begin{table*}[htbp]
\centering
\caption{Phase 2 Automated Evaluation: VLM Accuracy on Topological vs. Geometric Structures}
\label{tab:phase2_results}
\footnotesize
\resizebox{\textwidth}{!}{%
\begin{tabular}{llcccccc}
\toprule
& & \multicolumn{3}{c}{\textbf{Logic Diagrams (Topological)}} & \multicolumn{3}{c}{\textbf{Karnaugh Maps (Geometric)}} \\
\cmidrule(lr){3-5} \cmidrule(lr){6-8}
\textbf{Model} & \textbf{Condition} & \textbf{Pass@Any} & \textbf{Majority Vote} & \textbf{Strict (3/3)} & \textbf{Pass@Any} & \textbf{Majority Vote} & \textbf{Strict (3/3)} \\
\midrule

\multirow{2}{*}{\textbf{Gemini 2.5 Pro}}
& Clean      & \textbf{100\%} & \textbf{97.5\%} & 92.5\% & \textbf{100\%} & \textbf{90\%} & \textbf{90\%} \\
& Perturbed  & \textbf{82.5\%} & \textbf{76.2\%} & \textbf{71.2\%} & \textbf{100\%} & \textbf{100\%} & \textbf{70\%} \\
\midrule

\multirow{2}{*}{\textbf{Gemini 2.5 Flash}}
& Clean      & 96.2\% & 96.2\% & \textbf{95.0\%} & 86.7\% & 80.0\% & 73.3\% \\
& Perturbed  & 76.2\% & 70.0\% & 67.5\% & 86.7\% & 66.7\% & 53.3\% \\
\midrule

\multirow{2}{*}{\textbf{Claude 4.5 Haiku}}
& Clean      & 17.5\% & 15.0\% & 12.5\% & 26.7\% & 16.7\% & 10.0\% \\
& Perturbed  & 7.5\% & 6.2\% & 5.0\% & 13.3\% &  6.7\% &  3.3\% \\
\midrule

\multirow{2}{*}{\textbf{GPT-4o-mini}}
& Clean      & 10.0\% & 8.8\% & 6.2\% & 0.0\% & 0.0\% & 0.0\% \\
& Perturbed  & 5.0\% & 3.8\% & 2.5\% & 0.0\% & 0.0\% & 0.0\% \\

\bottomrule
\end{tabular}
}
\vspace{4pt}
\raggedright \footnotesize \textit{Note:} \textbf{Pass@Any} indicates $\geq 1$ correct inference across three trials. \textbf{Majority Vote} indicates $\geq 2$ correct inferences. \textbf{Strict} requires identical results for all three trials. Best results for both Clean and Perturbed are show in \textbf{bold}.
\end{table*}

\subsection{Phase 2: Automated Evaluation}
In Phase 2, we evaluated the models' structural vulnerabilities by automating the analysis of two distinct datasets: 80 Logic Diagrams (testing topological routing) and 30 Karnaugh Maps (testing  coordinate geometry). To account for stochasticity, we evaluated each image three times, recording Pass@Any (at least 1 of 3 passes correct), Majority Vote (at least 2 of 3 correct), and Strict Consistency (3 of 3 correct). We established a clean baseline with no perturbations, followed immediately by an evaluation using the perturbation suite of Warping, Noise, Lines, Photocopy. We specifically constrained Phase 2 to this ``full suite" approach due to practical resource limitations; conducting ablation studies on individual perturbations would require many millions of additional API tokens and substantial time. Testing this upper bound allowed us to observe whether the models could withstand maximum adversarial degradation.

The results reveal an appreciable disparity in geometric reasoning capabilities across the tested models. GPT-4o-mini experienced complete failure, achieving 0.0\% accuracy across all tiers on both the clean and perturbed datasets. This suggests a fundamental inability of the model's vision encoder to maintain the rigid 2D grid alignment required for K-Map minimization, possibly flattening the visual matrix into a linear sequence rather than processing it as a coordinate map. Claude 4.5 Haiku demonstrated limited baseline capability, achieving a 16.7\% Majority Vote on clean maps. Upon introducing adversarial perturbations, this performance degraded by more than half, dropping to a 6.7\% Majority Vote.

Gemini 2.5 Flash established the highest cognitive baseline, achieving an 80.0\% Majority Vote and 73.3\% Strict Consistency on clean images. However, the introduction of visual noise successfully degraded this performance, reducing the Majority Vote to 66.7\% and Strict Consistency to 53.3\%. While the model remained robust enough to solve the majority of the perturbed problems, the 20-point drop in Strict Consistency demonstrates that visual obfuscation successfully introduces instability into the model's reasoning paths. Ultimately, these results confirm that while perturbations act as a viable near-term stop-gap by degrading model reliability, robust models can still circumvent them, necessitating a structural shift in assessment design.

Gemini 2.5 Pro represented the most capable model in our evaluation. On the clean Karnaugh Map dataset, it achieved a 90.0\% Strict Consistency baseline. Curiously, while the introduction of visual noise paradoxically increased its Majority Vote accuracy to a perfect 100\%, its Strict Consistency simultaneously dropped to 70.0\%. This confirms that while the heaviest models possess excellent overall performance, the perturbations can still be applied successfully.

Conversely, the evaluation of the Logic Diagrams dataset focuses on the models' capacity for topological routing. As detailed in Table \ref{tab:phase2_results}, Gemini 2.5 Flash exhibited excellent baseline robustness, achieving a 95.0\% Strict Consistency on clean graphs. The introduction of the full perturbation suite successfully degraded this metric to 67.5\%. Gemini 2.5 Pro demonstrated similar foundational strength, achieving a 92.5\% Strict Consistency on clean graphs, which was successfully reduced to 71.2\%. Claude 4.5 Haiku and GPT-4o-mini demonstrated limited foundational capability on these topological tasks, achieving only single-digit Strict Consistency scores under perturbation. Given Gemini's ability to maintain a strong majority pass rate under heavy visual degradation we conjecture that sufficiently capable VLMs are inherently more resilient to noise when tracing continuous topological connections (wires) than when interpreting geometric alignments (K-Maps).

\section{Discussion}
\vspace{-5pt}
The results of Phase 1 (Table \ref{tab:unified_adversarial_results}) and Phase 2 (Table \ref{tab:phase2_results}) demonstrate that VLMs can possess substantially different effectiveness even on the unperturbed questions. Overall the Gemini family demonstrated good performance on both the original and perturbed questions, while models like GPT-4o-mini were largely incapable of solving even the original questions.

\subsection{Model Complexity and Paradoxical Improvement}
\vspace{-5pt}

In general, less complex models (Claude 4.5 Haiku, GPT-4o-mini) demonstrated proportionally more vulnerability to spatial perturbation, compared to larger frontier models (Gemini 2.5 Pro). However, multi-pass evaluation still revealed instability within these larger models. Across both the Phase 1 manual circuit evaluations and the Phase 2 Karnaugh Maps, Gemini 2.5 Pro occasionally exhibited ``positive perturbations'', where the introduction of noise shifted an incorrect baseline output to a correct one. In the Phase 2 K-Map dataset, perturbation increased the model's Majority Vote accuracy from 90.0\% to 100\%. We hypothesize that this is an instance of  ``paradoxical improvement'',  where a model performs better after the input data has been degraded, transformed, or otherwise made ``worse'' from a human perspective \parencite{zhai2023emergence,benzi1981mechanism}. We do note that the model's Strict Consistency simultaneously degraded from 90.0\% down to 70.0\%. This confirms that while the model was overall successful, the perturbations do have some limited effect.

\subsection{Efficacy of Perturbation Methods and Locations}
Across all model types, the ``Photocopy" perturbation proved more effective than the isolated Noise, Lines, and Warping (NLW) perturbations. In Phase 1, the photocopy method generated 15 successful negative output shifts, compared to only 7 from the standard suite. As the Photocopy perturbation on its own is generally less visually disruptive than the NLW, this was unexpected behavior as intuition would align with the more severe the perturbation, the worse the results. We hypothesize that this may be another case of paradoxical improvement.

We also note that isolating the location of the attack yielded counter-intuitive results. Perturbations applied solely to the question text were the least effective at forcing an incorrect output, while perturbations applied to the full combined image were occasionally less effective than perturbations applied exclusively to the figure region. We hypothesize that introducing noise to the text may trigger an ``error-correction" state within the VLM's language pathway, causing it to expend greater computational effort to align the text with the image, thereby inadvertently improving its performance. 
\vspace{-5pt}
\subsection{Human-Unrecognizable Inputs and the Limits of Obscurity}
While heuristic visual perturbations serve as a functional near-term deterrent for trivial plagiarism, initial threshold testing revealed a likely time limit on this methodology. Current vision transformer models demonstrated an alarming capacity to solve circuit problems that were degraded well beyond the threshold of human legibility (Figure \ref{fig:degraded}). Because VLMs can already parse certain heavily corrupted images better than humans, attempting to secure visual assessments purely through image degradation is likely a futile arms race. Long-term academic integrity must rely on alternative methods to secure assessments, rather than relying on visual obscurity.

\section{Limitations and Future Work}
While this work establishes a framework for evaluating VLM spatial reasoning, several limitations remain. First, Phase 2 utilized a combined ``full suite" perturbation approach due to the computational and token costs of testing at scale. Consequently, we did not perform an exhaustive ablation study isolating the efficacy of individual perturbations (e.g., affine warping versus pixel noise on Karnaugh Maps). We also caution against cross-phase comparison. Future work should conduct granular parameter tuning to identify the minimal perturbation required to induce model failure, as excess perturbation can increase model performance.

Second, model selection was intentionally constrained to represent the threat landscape most accessible to undergraduate students: free-tier web interfaces and optimized, low-cost ``lightweight flagship" APIs. We did not evaluate bleeding-edge frontier models such as the Gemini 3 or GPT-5 families. Given the observed capability gap between Gemini 2.5 Flash and Pro, higher-capacity models may possess greater resilience. Future research should benchmark these models to forecast the lifespan of visual obfuscation defenses.

Third, degraded problem sets may create accessibility challenges for students, particularly those with vision impairments or who rely on assistive technologies, potentially limiting the effectiveness of these perturbations as a learning aid. Instructor workload may also increase when selecting perturbations that degrade VLM performance while remaining accessible to students. This challenge will grow as VLMs evolve, making these perturbations a stopgap until VLM performance is empirically evaluated. Future work should examine perturbations alongside modified assessment designs to determine whether they are as effective as converting existing content to new assessment methods.

Finally, although outside the scope of our computational evaluation, the impact of adversarial perturbations on student behavior warrants investigation. Encountering degraded assignment images may reduce trust in VLM output, potentially encouraging students to verify the model's logic. A human-computer interaction study examining how visual perturbations affect trivial plagiarism and trust in model output may prove valuable.

\section{Conclusion}
This research demonstrates that modern Vision-Language Models present a significant challenge to academic integrity through ``trivial plagiarism,'' while also possessing exploitable weaknesses. Through a two-phase evaluation of introductory engineering and computing problems, we showed that heuristic image perturbations, such as affine warping and simulated photocopy degradation, can measurably degrade model accuracy in most configurations without destroying human interpretability. Testing also revealed appreciable capability differences between models, although even highly capable frontier models suffered measurable losses when evaluating perturbed problems.

Ultimately, computationally efficient visual perturbations provide a plausible near-term stopgap for educators seeking to deter low-effort academic dishonesty. However, as VLM capabilities increase, relying on ``security by visual obscurity" will likely become infeasible. To ensure the long-term validity of visual assessments, the academic community must shift toward assessment designs inherently resistant to VLM-based plagiarism.
\newline

\noindent \textbf{Data Availability Statement}: All materials used to generate the results in this paper are openly available at: \textbf{\small \url{https://github.com/christopherburger/AMVT}}




\printbibliography

\end{document}